%% file: _main.tex
\input{_constants}

\cameraready

\pdfoutput=1
\documentclass[10pt,twocolumn,letterpaper]{article}
\input{cvpr_header}
\begin{document}

\title{\paperTitle}
\author{\authorBlock}
\maketitle

\input{sections/00_abstract.tex}
\input{sections/01_intro.tex}
\input{sections/02_related.tex}
\input{sections/03_method.tex}
\input{sections/04_experiment.tex}
\input{sections/05_conclusion.tex}

{\small
\bibliographystyle{ieee_fullname}
\bibliography{references}
}


\end{document}

%% file: _constants.tex
\def\cvprPaperID{5546}
\def\confName{ICCV}
\def\confYear{2023}

\def\paperTitle{CTVIS: Consistent Training for Online Video Instance Segmentation}

\def\authorBlock{
    Kaining Ying$ ^{1,2}\thanks{KY (email: {$\tt kaining.ying.cv@gmail.com$}) and QZ contributed equally to this work. This work was done 
    when KY, QZ, WM, YZ were visiting 
     Zhejiang University.}  $     \quad
    Qing Zhong$ ^{4}\footnotemark[1]  $     \quad
    Weian Mao$^{4}$  \quad
    Zhenhua Wang$ ^{3}\thanks{Correponding authors.}  $  \quad
    Hao Chen$ ^{1}\footnotemark[2]  $ \quad
    \\ 
    Lin Yuanbo Wu$ ^{5}  $ \quad 
     Yifan Liu$ ^{4}  $ \quad
    Chengxiang Fan$ ^{1}  $ \quad
    Yunzhi Zhuge$ ^{4}  $ \quad
    Chunhua Shen$ ^{1}  $
    
    \\[.2cm]
    \normalsize $ ^1 $ Zhejiang University  \qquad
    \normalsize $ ^2 $ College of Computer Science and Technology, Zhejiang University of Technology \\
    \normalsize $ ^3 $ College of Information Engineering, Northwest A\&F University  \\
    \normalsize $ ^4 $ The University of Adelaide, Australia \qquad
    \normalsize $ ^5 $ Swansea University, UK \\
    \small \href{https://github.com/KainingYing/CTVIS}{https://github.com/KainingYing/CTVIS}
    
}

\newif\ifreview
\newif\ifarxiv
\newif\ifcamera\newcommand{\cameraready}{\cameratrue}
\newif\ifrebuttal

\newcommand{\blue}[1]{\textcolor{blue}{#1}}


%% file: cvpr_header.tex
\ifreview \usepackage[review]{cvpr} \fi
\ifarxiv \usepackage[pagenumbers]{cvpr} \fi
\ifrebuttal \usepackage[rebuttal]{cvpr} \fi
\ifcamera \usepackage{cvpr} \fi

\usepackage{graphicx}
\usepackage{amsmath}
\usepackage{amssymb}
\usepackage{booktabs}

\input{_macros}  

\usepackage{xr-hyper}

\makeatletter
\newcommand*{\addFileDependency}[1]{
  \typeout{(#1)}
  \@addtofilelist{#1}
  \IfFileExists{#1}{}{\typeout{No file #1.}}
}

\makeatother

\usepackage[pagebackref,breaklinks,colorlinks]{hyperref}
\usepackage[capitalize]{cleveref}
\crefname{section}{Sec.}{Secs.}
\crefname{table}{Table}{Tables}
\crefname{figure}{Fig.}{Figs.}

%% file: _macros.tex
\usepackage{times}
\usepackage{microtype}
\usepackage{epsfig}
\usepackage[table,xcdraw]{xcolor}
\usepackage{caption}
\usepackage{float}
\usepackage{placeins}
\usepackage{color, colortbl}
\usepackage{stfloats}
\usepackage{enumitem}
\usepackage{tabularx}
\usepackage{graphicx}
\usepackage{xstring}
\usepackage{multirow}
\usepackage{xspace}
\usepackage{url}
\usepackage{subcaption}
\usepackage{xcolor}
\usepackage{rotating}
\usepackage[hang,flushmargin]{footmisc}
\usepackage{bm}
\usepackage{amssymb}
\usepackage{pifont}

\ifcamera \usepackage[accsupp]{axessibility} \fi

% ...

\ifarxiv  \fi

\newcommand{\R}[1]{{%
    \textbf{%
        \ifstrequal{#1}{1}{\textcolor{red}{R#1}}{%
        \ifstrequal{#1}{2}{\textcolor{blue}{R#1}}{%
        \ifstrequal{#1}{3}{\textcolor{green!70!black}{R#1}}{%
        \ifstrequal{#1}{4}{\textcolor{teal}{R#1}}{%
                           \textcolor{cyan}{R#1}%
        }}}}%
    }%
}}

%% file: sections/00_abstract.tex
\begin{abstract}
    The discrimination of instance embeddings plays a vital role in associating instances across time for online video instance segmentation (VIS). Instance embedding learning is directly supervised by the contrastive loss computed upon the \textbf{contrastive items} (CIs), which are sets of anchor/positive/negative embeddings. Recent online VIS methods leverage CIs sourced from one reference frame only, which we argue is insufficient for learning highly discriminative embeddings. Intuitively, a possible strategy to enhance CIs is replicating the inference phase during training. To this end, we propose a simple yet effective training strategy, called \textbf{C}onsistent \textbf{T}raining for Online \textbf{VIS} (\textbf{CTVIS}), which devotes to aligning the training and inference pipelines in terms of building CIs. 
    Specifically, CTVIS constructs CIs by referring inference the momentum-averaged embedding and the memory bank storage mechanisms, and adding noise to the relevant embeddings. 
    Such an extension allows a reliable comparison between embeddings of current instances and the stable representations of historical instances, thereby conferring an advantage in modeling VIS challenges such as occlusion, re-identification, and deformation.  
    Empirically, CTVIS outstrips the SOTA VIS models by up to +5.0 points  on three VIS benchmarks, including YTVIS19 (55.1\% AP), YTVIS21 (50.1\% AP) and OVIS (35.5\% AP). 
    Furthermore, we find that pseudo-videos transformed from images can train robust models surpassing fully-supervised ones.
\end{abstract}

%% file: sections/01_intro.tex
\section{Introduction}
\label{sec:intro}

\begin{figure}[t]
    \centering
    \includegraphics[width=.85\columnwidth]{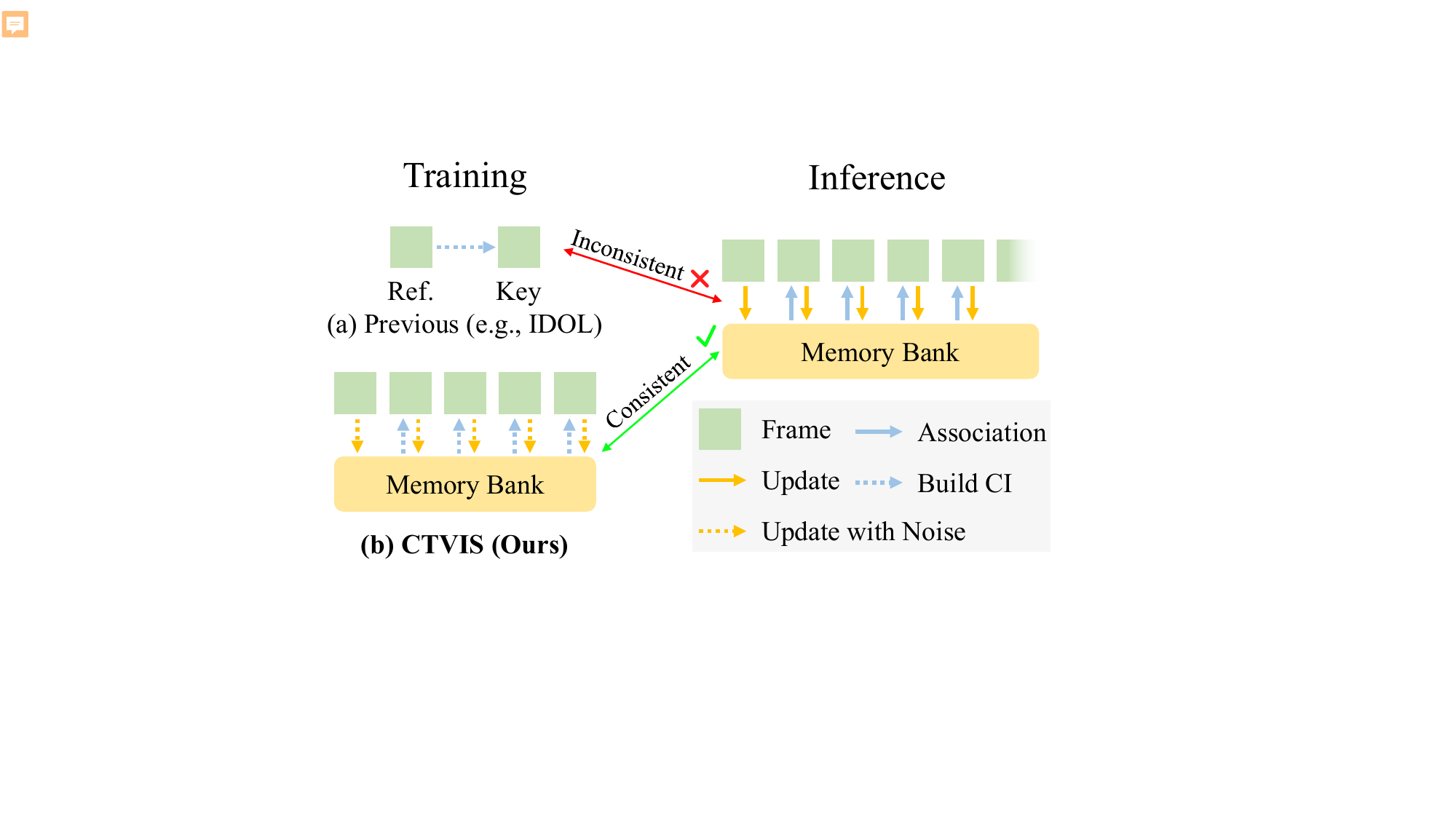}
    \vspace{-1mm}
    \caption{
    Comparison of inconsistent and consistent training (Ours). (a) Previous methods typically build contrastive items (CIs) and supervise the instance embeddings between key and reference frames. We call this paradigm inconsistent training, where the interaction with the long-term memory bank during training and the lack of modeling for long video in real inference scenarios is overlooked. (b) The purpose of \textbf{CTVIS} is to align the training and inference pipelines. Specifically, CTVIS constructs training stage CIs by leveraging the memory bank and incorporates noise during the memory bank updating to simulate real-world scenarios, such as ID switching, that can occur during inference.}
    \label{fig:teaser}
    \vspace{-2mm}
\end{figure}

Video instance segmentation is a joint vision task involving classifying, segmenting, and tracking interested instances across videos \cite{masktrackrcnn}. It is critical in many video-based applications, such as video surveillance, video editing, autonomous driving, augmented reality, \etc. Current mainstream VIS methods \cite{crossvis, masktrackrcnn, mask2formervis, idol, ifc, minvis, vita, vistr, seqformer} can be categorized into offline and online groups. The former \cite{ifc, mask2formervis, seqformer, vita, vistr} segments and classifies all video frames simultaneously and makes the instance association in a single step. The latter \cite{crossvis, idol, masktrackrcnn, minvis} takes as input a video in a frame-by-frame fashion, detecting and segmenting objects per frame while associating instances across time. In this paper, we focus on the online branch.

Online methods are typically built upon image-level instance segmentation models\cite{maskrcnn, mask2former, deformabledetr, fcos}. 
Several works \cite{masktrackrcnn, ovis, lsrf} utilize convolution-based instance segmentation models to segment each frame and associate instances by incorporating heuristic clues, such as mask-overlapping ratios and the similarity of appearance. 
However, these hand-designed approaches always fail to tackle complicated cases, which typically include severe target occlusion, deformation and re-identification. 
Recently, encouraged by the thriving of Transformer-based \cite{transformer} architectures in object detection and segmentation \cite{detr, deformabledetr, mask2former}, a bunch of query-based online frameworks have been proposed \cite{idol, minvis}, which take advantage of the temporal consistency of query embeddings and associate instances by linking corresponding query embeddings frame by frame. These advances boost the performance of online VIS models, which become de-facto leading VIS performance on most benchmarks (especially on challenging ones such as OVIS \cite{ovis}). 

Though the importance of the discrimination of query embeddings to associate instances has been nominated \cite{idol, minvis}, less research attention has been paid in this vein. 
MinVIS \cite{minvis} simply trains a single-frame segmentor, and the quality of its query embedding is hampered by the segmentor originally proposed for image-based instance segmentation. 
As shown in Figure~\ref{fig:teaser}(a), recent methods \cite{idol, stc} merely supervise instance embedding generation between two temporally adjacent frames with thein contrastive losses computed upon contrastive items.
Specifically, for each instance at the key frame, if the same instance appears on the reference frame, the embedding of it is selected as the anchor embedding $\mathbf{v}$. Meanwhile, its embedding in the reference frame is taken as the positive embedding $\mathbf{k}^+$, and the embeddings of other instances in the reference frame are used as the negative embeddings $\mathbf{k}^-$. In convention the set $\{\mathbf{v}, \mathbf{k}^+, \mathbf{k}^-\}$ is called \textbf{contrastive item} (CI). 
This training paradigm is \textit{inconsistent} with the inference (shown in the right of Figure~\ref{fig:teaser}), as it overlooks the interaction with the long-term memory bank to construct contrastive items and lacks modelling for long videos. To bridge this gap, we propose CTVIS (as shown in Figure~\ref{fig:teaser}(b)), which intuitively brings in useful tactics from inference, including memory bank, momentum-averaged (MA) embedding and noise training. 
Specifically, CTVIS samples several frames from a long video to form one training sample. Then we process each sample frame by frame, which can produce abundant CIs. Moreover, we sample momentum-averaged (MA) embeddings from the memory bank to create positive and negative embeddings. Furthermore, we introduce noise training for VIS, incorporating a few noises into the memory bank updating procedure to simulate the tracking failure scenarios during the inference process. 

We also consider the availability of large-scale training samples, which are especially expensive to annotate and maintain for VIS. To tackle this, we implement and test several goal-oriented augmentation methods (to align with the distribution of real data) to produce pseudo-videos. Different from the COCO joint training, we only use pseudo-videos to train VIS models. 

Without bells and whistles, CTVIS outperforms the state-of-the-art by large margins on all benchmark datasets, including YTVIS19 \cite{masktrackrcnn}, YTVIS22 \cite{masktrackrcnn}, and OVIS \cite{ovis}. Even trained with pseudo-videos only, CTVIS surpasses fully supervised VIS models \cite{idol, seqformer, vita}. Here we summarize our key contributions as


    

\begin{list}{\labelitemi}{\leftmargin=1em}
    \setlength{\topmargin}{0pt}
    \setlength{\itemsep}{0em}
    \setlength{\parskip}{0pt}
    \setlength{\parsep}{0pt}
    \item We propose a simple yet effective training framework (CTVIS) for online VIS. CTVIS promotes the discriminative ability of the instance embedding by 
    interacting with long-term memory banks to build CIs, 
    and by introducing noise into the memory bank updating procedure.

    \item We propose to create pseudo-VIS training samples by augmenting still images and their mask annotations. CTVIS models trained with pseudo-data only surpass their fully-supervised opponents already, which suggests that it is a desirable choice, especially when dense temporal mask annotations are limited.

    \item CTVIS achieves impressive performance on three public datasets. Meanwhile, extensive ablation validates the method's effectiveness.
    
  \end{list}

%% file: sections/02_related.tex
\section{Related Work}
\label{sec:related}

\noindent\textbf{Online VIS Method} \cite{masktrackrcnn, crossvis, minvis, idol, stc} typically builts upon image-level instance segmentation models \cite{maskrcnn, mask2former, deformabledetr, detr, isda}. MaskTrack R-CNN \cite{masktrackrcnn} extends Mask R-CNN \cite{maskrcnn} by incorporating an additional tracking head, which associates instances across videos using heuristic cues. CrossVIS \cite{crossvis} proposes to guide the segmentation of the current frame by the features extracted from previous frames. With the emergence of query-based instance segmentors \cite{deformabledetr, detr, mask2former}, matching with query embeddings instead of hand-designed rules boosts the performance of online VIS \cite{idol, minvis}. Utilizing the temporal consistency of intra-frame instance queries predicted by the image-level segmentor \cite{mask2former, deformabledetr}, MinVIS \cite{minvis} tracks instances by Hungarian matching of the corresponding queries frame by frame without video-based training. IDOL \cite{idol} supervises the matching between instances that appeared within two adjacent frames during training. During inference, IDOL maintains a memory bank to store instance momentum averaged embeddings detected from previous frames, which are employed to match with newly detected foreground instance embeddings. 
Concurrent work GenVIS \cite{genvis} applies a query-propagation framework to bridge the gap between training and inference in online or semi-online manners. 
Different from previous approaches, CTVIS aims to absorb ideas from the inference stage of online methods and learn more robust and discriminative instance embeddings during training. 


\noindent\textbf{Offline VIS Method} \cite{vistr, ifc, mask2formervis, vita, seqformer} takes as input the entire video and predicts masks for all frames in a single run. 
VisTR \cite{vistr} utilises clip-level instance features as input and predicts clip-level mask sequences in an end-to-end manner. Subsequently, several follow-up works, such as Mask2Former-VIS \cite{mask2formervis}, and SeqFormer \cite{seqformer}, exploit attention \cite{transformer} to process spatio-temporal features and directly predict instance mask sequences. To mitigate the memory consumption on extremely long videos, VITA \cite{vita} proposes to decode video object queries from sparse frame-level object tokens instead of dense spatio-temporal features.

\noindent\textbf{Discriminative Instance-Level Feature Learning.} The discrimination of instance embeddings plays a vital role in instance-level association tasks. Most works absorb the ideas from contrastive learning in self-supervised representation learning. IDOL \cite{idol} and QDTrack \cite{qdtrack} supervise the learning of contrastive instance representations between two adjacent frames. SimCLR \cite{simclr} argues that contrastive learning can benefit from larger batches. Inspired by this, CTVIS introduces long video training samples instead of key-reference image pairs, which leads to more robust instance embeddings. 

\noindent\textbf{VIS Model Training with Sparse Annotations.} Annotating masks for each object instance in every frame and linking them across the video is prohibitively expensive. 
Furthermore, recent works \cite{minvis, nguyen, qdtrack} suggest that the dense video annotations for VIS are unnecessary.  
MinVIS\cite{minvis} makes a per-frame image-level segmentation and associates the generated instance queries to obtain the video-level results. Since the training of the MinVIS model is agnostic to the temporal association of masks, it can benefit from the availability of large-scale datasets for image-level instance segmentation\cite{mscoco}. 
QDTrack \cite{qdtrack} learns compelling instance similarity using pairs of transformed views of images. 
MS COCO \cite{mscoco}, which contains abundant image-level mask annotations, is typically taken to supplement the training of models for VIS \cite{seqformer, idol, vita}. 
Following this, we propose to train VIS models with pseudo-videos generated by augmenting images instead of natural videos. We show that CTVIS models trained on pseudo-videos can surpass SOTA models \cite{seqformer, idol, vita, mask2formervis, masktrackrcnn, ifc} trained with densely annotated videos by clear margins. Different from techniques taking augmentation to enrich the training set \cite{detr, copypaste, qdtrack}, we use augmentation to create the set, which contains pseudo-videos and the associated mask annotations (as well as their spatio-temporal tracks). Moreover, we carefully design the video generation routines based on classical augmentation techniques (
i.e. \emph{rotation}, \emph{crop} and \emph{copy\&paste}), such that the pseudo-videos are realistic and can cover VIS challenges (including \emph{object occlusion}, \emph{fast-motion}, \emph{re-identification} and \emph{deformation}).

%% file: sections/03_method.tex
\section{Methods}
\label{sec:method}

CTVIS builds upon Mask2Former \cite{mask2former}, which is an effective image instance segmentation model (briefly reviewed in Section~\ref{sec:mask2former})\footnote{
Note that CTVIS can be easily combined with other query-based instance segmentation models \cite{idol, detr, deformabledetr} with minor modifications.}. Our CTVIS is motivated by the inference of typical online VIS methods introduced in Section~\ref{sec:inference}. 
Then we detail our consistent training method in Section~\ref{sec:ct}. Finally, Section~\ref{sec:pseudo} presents our goal-oriented pseudo-video generation technique for training VIS models with sparse image-level annotations.

\subsection{Brief Overview of Mask2Former} 
\label{sec:mask2former}
Mask2Former \cite{mask2former} composed of three main components: an \emph{image encoder} $\mathcal{E}$ (consist of a backbone and a pixel decoder), a \emph{transformer decoder} $\mathcal{T}$ and a \emph{prediction head} $\mathcal{P}$. Given an input image $I\in \mathbb{R}^{H \times W \times 3}$, $\mathcal{E}$ extracts a set of feature maps $\bm{F}=\mathcal{E}(I)$, where $\bm{F} = \{ F_0 \cdots F_{-1}\}$ is a sequence of multi-scale feature maps, and $F_{-1}$ is the final output of the $\mathcal{E}$ with $1/4$ resolution of $I$. The $N$ raw query embeddings $\hat{Q} \in \mathbb{R}^{N \times C}$ are learnable parameters, where $N$ is a large enough number of outputs and $C$ is the number of channels. Then, $\mathcal{T}$ takes both $\bm{F}$ and $\hat{Q}$ to iteratively refine query embeddings, and consequently outputs $Q \in \mathbb{R}^{N \times C}$. Finally, the prediction head outputs the segmentation masks $M$ and the classification scores $O$. For classification, $O=\mathcal{C}(Q) \in \mathbb{R}^{N \times K}$, where $K$ is the number of object categories. For  segmentation, the masks $M \in \mathbb{R}^{N \times H/4 \times W/4}$ are generated with $M = \sigma(Q \ast F_{-1})$, where $\ast$ denotes the convolution operation and $\sigma(\cdot)$ is the sigmoid function.

\noindent\textbf{Our Modification.} Because CTVIS employs instance embeddings to associate instances during inference, we add a  head (a few MLP layers) to compute the instance embeddings $E \in \mathbb{R}^{N \times C}$ based $Q$. 

\subsection{Inference of CTVIS}
\label{sec:inference}
CTVIS leverages Mask2Former\cite{mask2former} to process each frame 
and introduces an external memory bank\cite{idol, masktrackrcnn} to store the states of previously detected instances, including classification scores, segmentation masks and instance embeddings. 
To ease presentation, we assume that CTVIS has already processed $T$ frames out of an input video of $L$ frames, and there are $N$ predicted instances with $N$ instance embeddings $\bold{d}_i \in \mathbb{R}^C$ in the current frame. The memory bank stores for the previous $T$ frames $M$ detected instances, each of which has multiple temporal instance embeddings $\{ \bold{e}^t_j \in \mathbb{R}^C  \}^T_{t=1}$ and a momentum-averaged instance embedding $\hat{\bold{e}}_j^T$, which is computed according to the similarity-guided fusion \cite{sgf}: 
\begin{gather}
    \label{eq:sgf}
    \hat{\bold{e}}^T_j=(1-\beta^T) \hat{\bold{e}}^{T-1}_j+\beta^T \bold{e}^T_j \text {, } \\
    \beta^T=\max \left\{0, \frac{1}{T-1} \sum_{k=1}^{T-1} \Psi_d\left(e^T_j, e^{T-k}_j\right)\right\} , 
\end{gather}

\begin{figure*}[t]
    \vspace{-2mm}
    \centering
    \includegraphics[width=.95\textwidth]{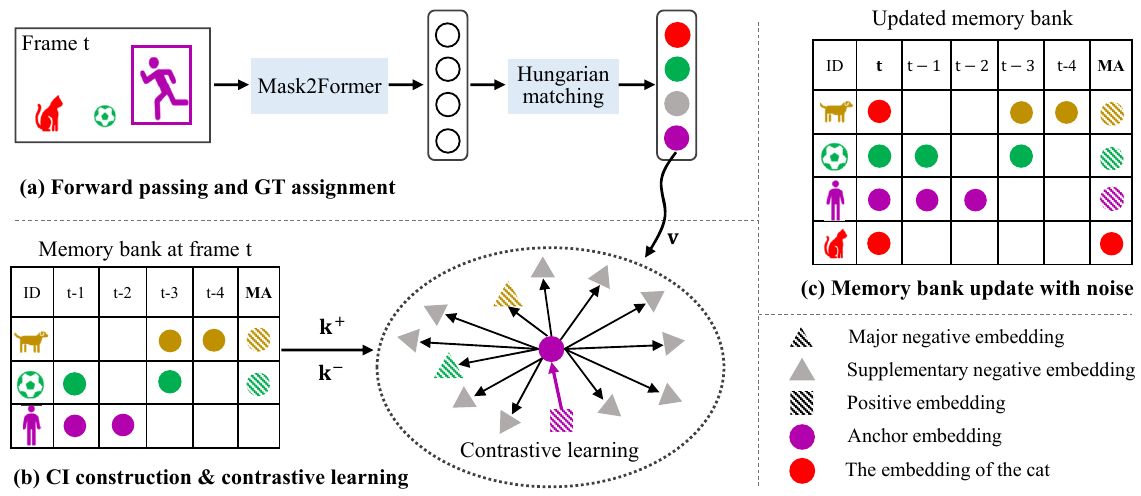}
    \vspace{-1mm}
    \caption{\textbf{Overview of the proposed CTVIS:} \textbf{a)} forward passing and GT assignment using Mask2Former and Hungarian matching; \textbf{b)} consistent training via building CIs with a memory bank. For simplicity, we only show the construction of CIs for the human instance (anchor) in the $t$-th frame of a training video. Through contrastive learning, positive embeddings are pulled close to the anchor embedding, while negative embeddings are pushed away from the anchor; \textbf{c)} Update the memory bank using the embeddings of frame $t$ with noise. } 
    \label{fig:main}
    \vspace{-3mm}
\end{figure*} 

\noindent where $\Psi_d$ denotes the cosine similarity. 
We refer the reader to \cite{sgf} for more details. Next, for each instance $i$ detected in the current frame, we compute its bi-softmax similarity \cite{qdtrack} with respect to the previously detected instance $j$ using

\begin{equation}
\label{equ:bio_softmax}
    f_{i,j}=
    0.5 \cdot \left[\frac{\exp \left(\hat{\mathbf{e}}_j^T \cdot \mathbf{d}_i\right)}{\sum_k \exp \left(\hat{\mathbf{e}}_k^T \cdot \mathbf{d}_i\right)}+\frac{\exp \left(\hat{\mathbf{e}}_j^T \cdot \mathbf{d}_i\right)}{\sum_{l} \exp \left(\hat{\mathbf{e}}_j^T \cdot \mathbf{d}_l\right)}\right] 
\end{equation}

Finally, we find the ``best''  
instance ID for $i$ with
\begin{equation}
\hat{j}=\arg \max f_{i,j}, \forall j \in\{1,2, \ldots, M\}.
\end{equation}
If $f_{i,\hat{j}} > 0.5$, we believe that newly detected instance $i$ and instance $\hat{j}$ in the memory bank correspond to the identical target. Otherwise, we initiate a new instance ID in the memory bank. When all frames are processed, the memory bank contains a certain number of instances, each of which takes a classification score list $\{c_i^t\}_{t=1}^{L}$ and a mask list $\{m_i^t\}_{t=1}^{L}$ (recall that $L$ denotes the number of frames). For each instance $i$, we calculate its video-level classification score by averaging the frame-level scores of the object. 

\subsection{Consistent Learning}
\label{sec:ct}

A reliable matching of instances (\ie using Equation~\eqref{equ:bio_softmax}) across time is required to track instances successfully. Hence the extraction of highly discriminative embeddings of objects is of great importance. 
We argue that the discrimination of instance embeddings extracted with recent models \cite{idol, stc} is still inadequate, especially for videos involving object-occlusion, shape-transformation and fast-motion. One main reason is that mainstream contrastive learning methods build CIs (\ie $\{\mathbf{v},\mathbf{k}^+,\mathbf{k}^-\}$) from the reference frame only, which results in the comparison of the anchor embedding against instantaneous instance embeddings in $\mathbf{k}^+$ and $\mathbf{k}^-$. Such embeddings are typically less discriminative and contain noise, which prevents training from learning robust representations. To address this, our CTVIS leverages a memory bank to store MA embeddings, thus supporting contrastive learning from more stable representations. Here our insight is to align the embedding comparison of training with that of inference (such that the two comparisons are consistent). Figure~\ref{fig:main} sketches our CTVIS, which processes the training video frame-by-frame. For an arbitrary frame $t$, CTVIS involves three steps: a) it first takes the Mask2Former and Hungarian matching to compute the instance embeddings, and to match them with GT (highlighted by red, green and purple); b) Then, it builds CIs using MA embeddings within the memory bank, and performs contrastive learning with CIs; and c) It updates the memory bank with noise (\eg the embedding of the \emph{cat} is deliberately added to the memory of the \emph{dog}), which serves the learning from the next frame.

\noindent\textbf{Forward passing and GT assignment.} As shown in Figure~\ref{fig:main}~(a), we first feed the current frame $t$ into Mask2Former to compute the embeddings for queries. Then we employ Hungarian matching to find an optimal match between the decoded instances and the ground truth (GT), such that each GT instance is assigned one instance embedding. Note that Hungarian matching relies on the costs calculated for all (\emph{Decoded-Instance}, \emph{GT-Instance}) pairs. Essentially, each cost measures the similarity between a pair of instances based on their labels and masks.

\noindent\textbf{Construct CIs.} 
After GT assignment, we build the contrastive items for each GT instance using a memory bank. The memory bank stores all detected instances of previous $t-1$ frames, each associated with 1) a series of instance embeddings extracted at different times, and 2) its MA embedding computed by Equation~\eqref{eq:sgf}. 
In order to prepare the CIs $\{\mathbf{v}, \mathbf{k}^+, \mathbf{k}^-\}$ for instance $i$ (termed as the \emph{anchor}, \eg the person in Figure~\ref{fig:main}~(a)) at the $t$-th frame, the instance embedding extracted from this frame is used as the anchor embedding $v$.
For the positive embedding, we pick from the memory bank the MA embedding of instance $i$.
The negative embeddings $\mathbf{k}^-$ include the major negative embeddings and the supplementary negative embeddings. We use the MA embeddings of other instances in the memory bank as the major negative embeddings. We also sample the background query embeddings of previous $t - 1$ frames to form the supplement negative embeddings. Taking as inputs the created CIs, we compute the contrastive loss with
\begin{equation}
\label{eq:loss_embed}
\begin{aligned}
    \mathcal{L}_{\text {emb}} & =-\log \frac{\exp \left(\mathbf{v} \cdot \mathbf{k}^{+}\right)}{\exp \left(\mathbf{v} \cdot \mathbf{k}^{+}\right)+\sum\nolimits_{\mathbf{k}^{-}} \exp \left(\mathbf{v} \cdot \mathbf{k}^{-}\right)} \\
    & =\log \left[1+\sum\nolimits_{\mathbf{k}^{-}} \exp \left(\mathbf{v} \cdot \mathbf{k}^{-}-\mathbf{v} \cdot \mathbf{k}^{+}\right)\right].
\end{aligned}
\end{equation}
As shown in Figure~\ref{fig:main} (c), training with $\mathcal{L}_{\text {emb}}$ pulls the embeddings of positive instances close to the anchor embedding, while pushing the negative embeddings away from it.

\noindent\textbf{Update memory bank.} After computing the $\mathcal{L}_{\text{emb}}$ for each instance in frame $t$, we need to update the memory bank, such that the updated version can be taken to build CIs for frame $t+1$.
Unlike the inference stage, for training we can get the ground truth ID of each instance so as to update the memory bank with their embeddings extracted from frame $t$.
In comparison, inference can fail to track instances across time (\ie the ID switch issue), especially for complicated scenarios. To alleviate this, we introduce noise to the update of the memory bank, which compels the contrastive learning to tackle the switch of instance IDs.
Specifically, each disappeared instance (\eg the dog) in frame $t$ will have a little chance to receive an embedding of other instances (\eg the cat, which is randomly picked from all available instances) in the same frame, which is called the \emph{noise}. 
%
%
If the generated random value exceeds a threshold (\eg 0.05), as illustrated in Figure~\ref{fig:main}~(c), we use the noise as the embedding of the disappeared instance at frame $t$. Finally, the MA embeddings are updated for all instances using Equation~\eqref{eq:sgf}. Due to the low similarity between the disappeared instance and the noise, such an update has quite a limited impact on the MA embedding of the instance, which is reidentified later. Indeed, training with noise is able to reduce the chance of ID switch, as demonstrated by the fish example in Figure~\ref{fig:video}. 

\noindent\textbf{Loss.} After processing all frames, The $\mathcal{L}_{\text {emb}}$ values of all CIs are averaged to obtain $L_{\text {emb}}$.
The total training loss is
\begin{equation}
L_{\text{total}} = \lambda_{\text{emb}}L_{\text{emb}} + \lambda_{\text{cls}} L_{\text{cls}} + \lambda_{\text{ce}} L_{\text {ce}} + \lambda_{\text{dice}} L_{\text{dice}},
\end{equation}
where $\lambda$ denotes loss weight. $L_{\text {cls}}$, $L_{\text {ce}}$ and $L_{\text {dice}}$ supervise the per-frame segmentation as suggested in \cite{mask2former}.

\vspace{-3mm}
\subsection{Learning from Sparse Annotation}
\label{sec:pseudo}

\begin{figure}
    \centering
    \includegraphics[width=0.9\columnwidth]{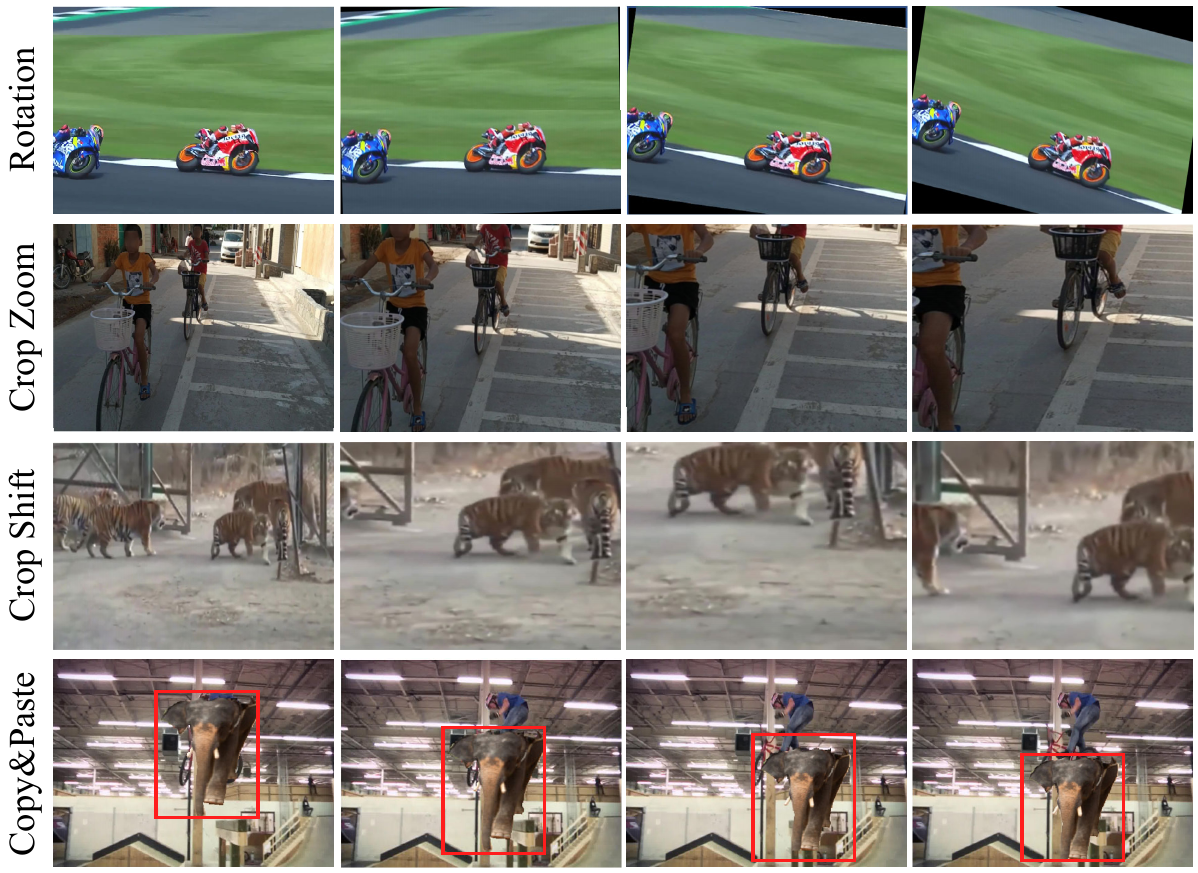}
    \vspace{-2mm}
    \caption{Generating pseudo-videos by augmenting images.}
    \label{fig:augs}
    \vspace{-4mm}
\end{figure} 

We now elaborate on our pseudo-video and mask generation technique, which enables the training of VIS models when only sparse annotations (\eg image data) are available. We take a few widely applied image-augmentation methods, including \emph{random rotation}, \emph{random crop} and \emph{copy\&paste} on source image to create pseudo-videos and the associated instance masks. Note that the pseudo-videos are created by no means to approximate real ones. Instead, they are taken to mimic the movement of targets in reality. 

\noindent\textbf{Rotation.} 
As shown in the first row of Figure~\ref{fig:augs}, the rotation augmentation rotates the source images with several random angles (e.g., $ [-15, 15]$ ) to introduce subtle changes between frames of the pseudo-videos. 

\noindent\textbf{Crop.} 
The rotation augmentation cannot alter the shapes and magnitudes of instances. However, instances deform or/and enter/exit the visible field due to the movement introduced either by the target or the camera. To address this, we apply random crop augmentation to the image, which allows the generated videos to mimic the zooming in/out effect of the camera lens and the shifting of targets. The second and the third rows of Figure~\ref{fig:augs} present two examples of \emph{crop-zoom} and \emph{crop-shift}, respectively. The pseudo-videos generated by such augmentations cover a large proportion of targets' movements.

\noindent\textbf{Copy and Paste.} 
As mentioned earlier, the trajectories of instances in pseudo-videos created by the augmentations share the identical motion direction. To incorporate the relative motion between instances, we also employ the \emph{copy\&paste} augmentation\cite{copypaste}, which copies the instances from another image in the dataset and pastes them into  random locations within the source image. Note that the pasting positions of an instance are typically different across time, which brings the relative motion between different instances (as shown in the fourth row of Figure~\ref{fig:augs}).



%% file: sections/04_experiment.tex
\section{Experiment}
\label{sec:experiment}

\noindent\textbf{Datasets.} 
The proposed methods are evaluated on three VIS benchmarks: YTVIS19\cite{masktrackrcnn}, YTVIS21\cite{masktrackrcnn} and OVIS\cite{ovis}.


\begin{table*}[!htbp]
\vspace{-2mm}
  \centering
    \resizebox{0.95\textwidth}{!}{%
    \begin{tabular}{c|c|c|ccccc|ccccc|ccccc}
    \toprule
    \multicolumn{2}{c|}{\multirow{2}[1]{*}{Methods}} & \multirow{2}[1]{*}{\shortstack{Params.}} & \multicolumn{5}{c|}{YTVIS19\cite{masktrackrcnn}}                               & \multicolumn{5}{c|}{YTVIS21\cite{masktrackrcnn}}                               & \multicolumn{5}{c}{OVIS\cite{ovis}} \\
    \multicolumn{2}{c|}{}     &             & AP          & AP$_{\mathtt{50}}$        & AP$_{\mathtt{75}}$        & AR$_{\mathtt{1}}$         & AR$_{\mathtt{10}}$        & AP          & AP$_{\mathtt{50}}$        & AP$_{\mathtt{75}}$        & AR$_{\mathtt{1}}$         & AR$_{\mathtt{10}}$        & AP          & AP$_{\mathtt{50}}$        & AP$_{\mathtt{75}}$        & AR$_{\mathtt{1}}$         & AR$_{\mathtt{10}}$ \\
    \midrule
    \multirow{11}[2]{*}{\begin{sideways}ResNet-50\cite{resnet}\end{sideways}} & MaskTrack R-CNN\cite{masktrackrcnn} & -           & 30.3        & 51.1        & 32.6        & 31          & 35.5        & 28.6        & 48.9        & 29.6        & 26.5        & 33.8        & 10.8        & 25.3        & 8.5         & 7.9         & 14.9 \\
                & SipMask\cite{sipmask}     & -           & 33.7        & 54.1        & 35.8        & 35.4        & 40.1        & 31.7        & 52.5        & 34          & 30.8        & 37.8        & 10.2        & 24.7        & 7.8         & 7.9         & 15.8 \\
                & CrossVIS\cite{crossvis}    & -           & 36.3        & 56.38       & 38.9        & 35.6        & 40.7        & 34.2        & 54.4        & 37.9        & 30.4        & 38.2        & 14.9        & 32.7        & 12.1        & 10.3        & 19.8 \\
                & IFC\cite{ifc}         & -           & 41.2        & 65.1        & 44.6        & 42.3        & 49.6        & 35.2        & 55.9        & 37.7        & 32.6        & 42.9        & 13.1        & 27.8        & 11.6        & 9.4         & 23.9 \\
                & Mask2Former-VIS\cite{mask2formervis} & 44          & 46.4        & 68          & 50          & -           & -           & 40.6        & 60.9        & 41.8        & -           & -           & 17.3        & 37.3        & 15.1        & 10.5        & 23.5 \\
                & TeViT\cite{tevit}       & -           & 46.6        & 71.3        & 51.6        & 44.9        & 54.3        & 37.9        & 61.2        & 42.1        & 35.1        & 44.6        & 17.4        & 34.9        & 15          & 11.2        & 21.8 \\
                & SeqFormer\cite{seqformer}   & 48        & 47.4        & 69.8        & 51.8        & 45.5        & 54.8        & 40.5        & 62.4        & 43.7        & 36.1        & 48.1        & 15.1        & 31.9        & 13.8        & 10.4        & 27.1 \\
                & MinVIS\cite{minvis}      & 44          & 47.4        & 69          & 52.1        & 45.7        & 55.7        & 44.2        & 66          & 48.1        & 39.2        & 51.7        & 25          & 45.5        & 24          & 13.9        & 29.7 \\
                & IDOL\cite{idol}        & \textbf{43}        & 49.5        & \underline{74}          & 52.9        & 47.7        & 58.7        & 43.9        & \underline{68}          & \underline{49.6}        & 38          & 50.9        & \underline{30.2}        & \underline{51.3}        & \underline{30}          & \underline{15}          & \underline{37.5} \\
                & VITA\cite{vita}        & 57        & \underline{49.8}        & 72.6        & \underline{54.5}        & \underline{49.4}        & \underline{61}          & \underline{45.7}        & 67.4        & 49.5        & \underline{40.9}        & \underline{53.6}        & 19.6        & 41.2        & 17.4        & 11.7        & 26 \\
                & \textbf{CTVIS (Ours)} & \underline{44}          & \textbf{55.1} & \textbf{78.2} & \textbf{59.1} & \textbf{51.9} & \textbf{63.2} & \textbf{50.1} & \textbf{73.7} & \textbf{54.7} & \textbf{41.8} & \textbf{59.5} & \textbf{35.5} & \textbf{60.8} & \textbf{34.9} & \textbf{16.1} & \textbf{41.9} \\
    \midrule
    \multirow{6}[1]{*}{\begin{sideways}Swin-L \cite{swin}\end{sideways}} 
    & SeqFormer\cite{seqformer}   & 219       & 59.3        & 82.1        & 66.4        & 51.7        & 64.6        & 51.8        & 74.6        & 58.2        & 42.8        & 58.1        & -           & -           & -           & -           & - \\
                & Mask2Former-VIS\cite{mask2formervis} & 216       & 60.4        & 84.4        & 67          & -           & -           & 52.6        & 76.4        & 57.2        & -           & -           & 25.8        & 46.5        & 24.4        & 13.7        & 32.2 \\
                & MinVIS\cite{minvis}      & 216         & 61.6        & 83.3        & 68.6        & 54.8        & 66.6        & 55.3        & 76.6        & 62          & 45.9        & 60.8        & 39.4        & 61.5        & 41.3        & \underline{18.1}        & 43.3 \\
                & VITA\cite{vita}        & 229       & 63          & 86.9        & 67.9        & \underline{56.3}        & 68.1        & \underline{57.5}        & 80.6        & 61          & \underline{47.7}        & \underline{62.6}        & 27.7        & 51.9        & 24.9        & 14.9        & 33 \\
                & IDOL\cite{idol}        & \textbf{213}       & \underline{64.3}        & \underline{87.5}        & \underline{71}          & 55.5        & \underline{69.1}        & 56.1        & \underline{80.8}        & \underline{63.5}        & 45          & 60.1        & \underline{42.6}        & \underline{65.7}        & \underline{45.2}        & 17.9        & \underline{49.6} \\
                & \textbf{CTVIS (Ours)} & \underline{216}         & \textbf{65.6} & \textbf{87.7} & \textbf{72.2} & \textbf{56.5} & \textbf{70.4} & \textbf{61.2} & \textbf{84} & \textbf{68.8} & \textbf{48} & \textbf{65.8} & \textbf{46.9} & \textbf{71.5} & \textbf{47.5} & \textbf{19.1} & \textbf{52.1} \\
    \bottomrule
    \end{tabular}%
    }
    \vspace{-1mm}
    \caption{Compare CTVIS with SOTA methods. The best and second best are highlighted by \textbf{bold} and \underline{underlined} numbers, respectively.
    }
    \label{tab:main}%
    \vspace{-3mm}
\end{table*}%

\noindent\textbf{Metrics.}
Following prior studies \cite{idol, minvis, mask2formervis, masktrackrcnn, crossvis, ifc, seqformer, vita}, we use Average Precision (AP) and Average Recall (AR) as the evaluation metrics. 

\noindent\textbf{Implementation Details.}
For the hyper-parameters of Mask2Former\cite{mask2former}, we just use its officially released version. The number of layers of the instance embedding head is 3. All models are initialized with parameters pre-trained on COCO \cite{mscoco}, and then they are trained on 8 NVIDIA A100 GPUs. Following prior works \cite{seqformer, vita, genvis},  we use the COCO joint training (CJT) setting to train our models unless otherwise specified. We set the lengths of training videos as 8 and 10 for YTVIS19\&21 and OVIS, respectively. For data augmentation, we use clip-level random crop and flip. During the training phase, we resize the input frames so that the shortest side is at least 320 and at most 640p, while the longest side is at most 768p. During inference, the input frames are downsampled to 480p. We set $\lambda_{\text{emb}}$, $\lambda_{\text{cls}}$, $\lambda_{\text{ce}}$, $\lambda_{\text{dice}}$ as 2.0, 2.0, 5.0 and 5.0, respectively. The mini-batch size is 16 and the maximum training iterations is 16,000. The initial learning rate is 0.0001 and decays at 6,000 and 12,000 iterations, respectively. 

\subsection{Main Result}
As shown in Table~\ref{tab:main}, we compare CTVIS against SOTA methods \cite{masktrackrcnn, sipmask, crossvis, ifc, mask2formervis, tevit, seqformer,minvis,idol,vita}, respectively using ResNet-50 \cite{resnet} and Swin-L \cite{swin} as the backbone on three benchmarks.

\noindent\textbf{YTVIS19 \& YTVIS21.} consist of relatively simple videos with short durations. Thanks to the introduced consistent learning paradigm and the extracted discriminative embeddings, CTVIS outperforms recent best methods on AP by $\sim5\%$ with ResNet-50 on both benchmarks. With the stronger backbone Swin-L, CTVIS surpasses the second best by $3.7\%$ on YTVIS21. Compared with IDOL\cite{idol}, CTVIS considerably improves the performance in terms of all metrics with tolerable parameter overheads. 

\noindent\textbf{OVIS.} This dataset contains longer videos and more intricate contents, on which online methods \cite{idol, minvis} perform much better than offline models \cite{vita, mask2former, seqformer}. 
Thanks to the effective embedding learning with long video samples, CTVIS gains $5.3$ and $4.3$ points in terms of AP, taking as inputs ResNet-50 and Swin-L, respectively. To summarize, CTVIS is highly competitive on benchmarks with varying complexities.

\subsection{Ablation Study}
We conduct extensive ablation to verify the effectiveness of CTVIS. Unless specified otherwise, we take the ResNet-50 as the backbone and train models under the CJT setting. Here we report AP$^{\mathtt{YV19}}$ and AP$^{\mathtt{OVIS}}$ on YTVIS19 and OVIS.

\noindent\textbf{Do improvements mainly come from better image-level instance segmentation models?} 
The answer is no. We validate this in Table~\ref{tab:detector}: 
1) Compared with IDOL with Deformable DETR, IDOL with Mask2Former is 1.0 and 1.5 points higher, suggesting the influence of a better detector is not that significant; 
2) Since our CTVIS is not restricted to a specific network, we implement Deformable DETR with CTVIS, which brings 4.2  and 3.6 points of AP gains. Similarly, CTVIS on Mask2Former also boosts the results by 3.9 and 3.8 points, which indicates that the improvements mainly come from our proposed CTVIS. 

\begin{table}[t]
\vspace{-2mm}
  \centering
  \resizebox{0.9\columnwidth}{!}{%
    \begin{tabular}{c|ll|ll}
    \toprule
    \multirow{2}[2]{*}{Methods} & \multicolumn{2}{c|}{Deformable DETR$^*$ \cite{idol}} & \multicolumn{2}{c}{Mask2Former \cite{mask2former}} \\
                           & AP$^{\mathtt{YV19}}$     & AP$^{\mathtt{OVIS}}$        & AP$^{\mathtt{YV19}}$     & AP$^{\mathtt{OVIS}}$ \\
    \midrule
    IDOL\cite{idol}        & 49.5        & 30.2        & 51.2        & 31.7 \\
    CTVIS                  &  \textbf{53.7} \blue{(+4.2)}        &  \textbf{33.8} \blue{(+3.6)}        &  \textbf{55.1} \blue{(+3.9)}        &  \textbf{35.5} \blue{(+3.8)} \\
    \bottomrule
  \end{tabular}%
  }
  \vspace{-1mm}
  \caption{Comparison of different instance segmentation methods with IDOL and CTVIS, respectively. 
  Deformable DETR$^*$ is extended to instance segmentation as suggested in \cite{idol}. }  
  \label{tab:detector}%
  \vspace{-1mm}
\end{table}%

\begin{figure}[t]
    \centering
    \includegraphics[width=0.9\columnwidth]{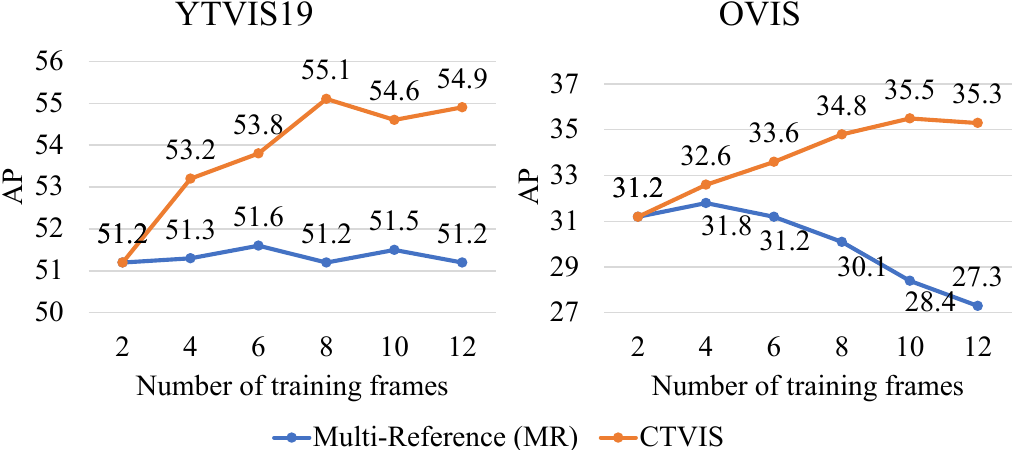}
    \vspace{-1mm}
    \caption{Ablation on the number of training frames. Multi-Reference extends IDOL by using multiple reference frames and the stronger Mask2Former as the segmentor.}
    \vspace{-3mm}
    \label{fig:ablate_num_frame}
\end{figure} 

\noindent\textbf{Long-video training.}
To verify the effectiveness of long-video training, we ablate the number of frames of each video used for training. For a fair comparison, we extend IDOL\cite{idol} to a multiple references (MR) version, by replacing its segmentor with the stronger Mssk2Former and using multiple reference frames.  Figure~\ref{fig:ablate_num_frame} shows the results. Thanks to the CI construction method employed by CTVIS, the performance has seen a dynamic increase by using more frames (peaked at 8 and 10 frames). In comparison, MR cannot benefit from long-video training and even degrades on OVIS. Hence we conclude that the performance of CTVIS stems from the effective video-level embedding learning (for tracking), rather than training an enhanced instance segmentor with larger batch sizes (more images per batch).

\noindent\textbf{Components of CTVIS.} First, removing all components of CTVIS sets a baseline, which utilizes a single reference to learn embeddings in a frame-by-frame way. As shown in Table~\ref{tab:ctvis}, the baseline gets 51.6 and 32.6 on YTVIS19 and OVIS. Based on this baseline, we gradually add CTVIS components: 1) We take the latest embedding of each instance to build CIs (instead of MA embeddings), which improves AP$^{\mathtt{YV19}}$ and AP$^{\mathtt{OVIS}}$ to 52.1 and 33.3. This suggests that the sampling domain CIs do indeed influence the instance embedding learning; 2) When MA is incorporated, the results see salient increases (52.1 \vs 54.2 and 33.3 \vs 34.9), which indicates that our CI-building method renders the embedding learning more stable and consistent; 3) When incorporating noise in the memory bank, which is designed to alleviate the ID switch issue, the performance sees non-trivial increases (0.9 and 0.6 on two datasets). Put all components together, CTVIS obtains remarkable results on both datasets and outperforms the strong baseline by 3.5 and 2.9 points, which validates the significance of the temporal alignment between training and inference pipelines, at least for VIS. 

\noindent\textbf{Sampling of $\mathbf{k}^-$.} 
We test different ways of building the negative embeddings $\mathbf{k}^-$. Table~\ref{tab:negative} presents four configurations and the corresponding results. Recall that the supplementary negative embeddings represent the background, and training with such negative samples only corrupts the performance (the 1st row). On the other hand, using major negative samples only gives decent results. A conjunctive usage of both negative-sampling types improves the performance significantly. In this line, we further consider sampling supplementary negative instances from either the local (sampled from the preceding frame only) or global domain (sampled from all previous frames). We found that the local setting gives the best results. This is probably because the model only needs to check the background in the local domain during inference. Hereafter we simply use the local setting.

\begin{table}[t]
\vspace{-2mm}
  \centering
  \resizebox{0.85\columnwidth}{!}{%
    \begin{tabular}{ccc|cc}
    \toprule
    Memory bank & Momentum & Noise & AP$^{\mathtt{YV19}}$ & AP$^{\mathtt{OVIS}}$     \\
    \midrule
                     &                  &                  & 51.6  & 32.6                     \\
    $\checkmark$     &                  &                  & 52.1  & 33.3                     \\
    $\checkmark$     & $\checkmark$     &                  & 54.2  & 34.9                     \\
    $\checkmark$     & $\checkmark$     & $\checkmark$     & \textbf{55.1}  & \textbf{35.5}   \\
    
    \bottomrule
    \end{tabular}%
  }
  \vspace{-1mm}
  \caption{Effectiveness of different CTVIS components.}
  \label{tab:ctvis}
\end{table}%

\begin{table}[t]
\vspace{-2mm}
  \centering
  \resizebox{0.7\columnwidth}{!}{%
    \begin{tabular}{cc|cc}
    \toprule
    Major       & Supplementary          & AP$^{\mathtt{YV19}}$ & AP$^{\mathtt{OVIS}}$ \\
    \midrule
                & $\checkmark$           &   16.5      &  0.5  \\
    $\checkmark$           &             &   50.8	   &  31.6 \\
    $\checkmark$           & global      &   54.6      &  33.4 \\
    $\checkmark$           & local       &   \textbf{55.1}      &  \textbf{35.5} \\
    \bottomrule
    \end{tabular}%
  }
  \vspace{-1mm}
  \caption{Ablate the sampling strategy of negative embeddings.} 
  \vspace{-3mm}
  \label{tab:negative}%
\end{table}%

\subsection{Pseudo Video as Training Example}
\label{sec:exp_pseudo}
We train VIS models on pseudo-videos, which are created with COCO images and the method described in Section~\ref{sec:pseudo}. Since COCO classes do not match that of VIS datasets, we only adopt the overlapping categories for training. For evaluation, we sample 421 and 140 videos with overlapping categories from the train sets of YTVIS21 and OVIS train sets, respectively. For more dataset information, please refer to the supplementary material. Specially, we denote the sampled version of YTVIS21 and OVIS as YTVIS21$^*$ and  OVIS$^*$. We use Swin-L as the backbone, and investigate the impacts of augmentation techniques in terms of generating pseudo-video datasets for training. Here \emph{rotation} is taken as the baseline. As shown in Table~\ref{tab:augs}, both \emph{crop} and \emph{copy\&paste} bring gains on both datasets over the baseline. Because YTVIS21 is relatively simple, \emph{crop} and \emph{copy\&paste} only improve the results by $0.2$ and $0.5$, respectively. However, for the complicated OVIS, they offer much larger performance gains, \ie $1.3$ and $2.0$ on two datasets, which suggests that pseudo videos generated with stronger augmentations are especially suitable to tackle complicated VIS tasks. We also train VITA and IDOL models using the generated pseudo-samples. Again, CTVIS surpasses them by clear margins, as that demonstrated in Table~\ref{tab:pseudo_sota}.

\subsection{Training with Limited Supervision}
Following MinVIS\cite{minvis}, we train CTVIS and MinVIS models on only a proportion ($\%$) of VIS training set. Specifically, we sample 1\%, 5\%, 10\%, and 100\% frames respectively from the training set to create pseudo videos for training. As shown in Table~\ref{tab:st_img}, with a 5\% proportion, CTVIS outperforms MinVIS with 100\% samples on all datasets.  More importantly, CTVIS trained with pseudo videos, which are created from 100\% frame samples, even surpasses the fully supervised competitors, and achieves close performance compared with CTVIS learned from genuine videos. 

\begin{table}[t]
  \centering
  \resizebox{0.8\columnwidth}{!}{%
    \begin{tabular}{ccc|cc}
    \toprule
        Rotation  &Crop  & Copy\&Paste       & AP$^{\mathtt{YV21^*}}$ & AP$^{\mathtt{OVIS^*}}$ \\
    \midrule
    $\checkmark$ &  &                                      & 48.5             & 27.3          \\
    $\checkmark$ & $\checkmark$  &                         & 48.7             & 28.6           \\
    $\checkmark$ &  & $\checkmark$                         & 49               & 29.3    \\
    $\checkmark$ & $\checkmark$ & $\checkmark$             & \textbf{49.7}             &\textbf{30.5}    \\
    \bottomrule
  \end{tabular}%
  }
  \vspace{-1mm}
  \caption{Influence of augmentations on producing pseudo-videos.}
  \label{tab:augs}%

\end{table}%

\begin{table}[t]
\vspace{-1mm}
\centering
\resizebox{0.8\columnwidth}{!}{%
\begin{tabular}{c|c|cc}
\toprule
    Methods &   Supervision &  AP$^{\mathtt{YV21^*}}$ & AP$^{\mathtt{OVIS^*}}$ \\ 
    \midrule
    MinVIS \cite{minvis}  &  Image  & 43.9             & 24.4 \\
    VITA   \cite{vita}    &  Pseudo video & 44.4             & 19.1 \\
    IDOL   \cite{idol}    &  Pseudo image pair& 47.8             & 27.8 \\
    CTVIS                 &  Pseudo video & \textbf{49.7}             & \textbf{30.5} \\
    \bottomrule
\end{tabular}%
}
\vspace{-1mm}
\caption{Compare with SOTA models trained with pseudo-samples, which are generated based on COCO images.}
\label{tab:pseudo_sota}
\vspace{-3mm}
\end{table}

\begin{figure*}[ht]
    \vspace{-2mm}
    \centering
    \includegraphics[width=.9\textwidth]{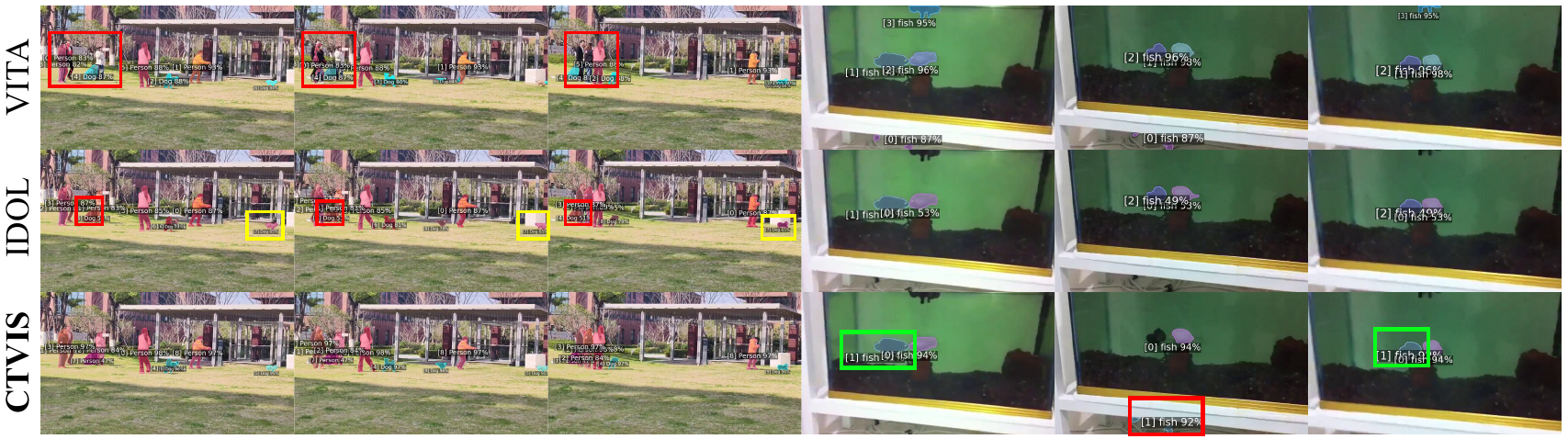}
    \vspace{-1mm}
    \caption{Visualize VIS results obtained by VITA \cite{vita}, IDOL \cite{idol} and CTVIS. These examples show performance under heavy occlusion (the left example), sudden lighting-condition change (the right example) and disturbance of targets of the same category (the right example). Here, red boxes highlight inferior segmentations, and yellow ones mark incorrect IDs.}
    \vspace{-2mm}
    \label{fig:video}
\end{figure*} 

\begin{table}[t]
    \centering
    \footnotesize
    \begin{tabular}{c|c|ccc}
    \toprule
    Methods                  & Training                 & AP$^{\mathtt{YV19}}$ & AP$^{\mathtt{YV21}}$ & AP$^{\mathtt{OVIS}}$ \\
    \midrule
    VITA \cite{vita}        & \multirow{3}{*}{Full} & 63           & 57.5         & 27.7      \\
    IDOL \cite{idol}        &                       & 64.3         & 56.1         & 42.6      \\
    \textbf{CTVIS (Ours)}   &                       & \textbf{65.6}         & \textbf{61.2}         & \textbf{46.9}      \\
    \midrule
    \multirow{4}{*}{MinVIS \cite{minvis}} 
                            & 1\%                   & 59           & 52.9         & 31.7 \\
                            & 5\%                   & 59.3         & 54.3         & 35.7 \\
                            & 10\%                  & 61           & 54.9         & 37.2 \\
                            & 100\%                 & \textbf{61.6}         & \textbf{55.3}         & \textbf{39.4} \\
    \midrule
    \multirow{4}{*}{\textbf{CTVIS (Ours)}}  
                            & 1\%                   & 62.4         & 57.8        & 36.2 \\
                            & 5\%                   & 63.4         & 59.4         & 41.9 \\
                            & 10\%                  & 64.2         & 60.0         & 42.1 \\
                            & 100\%                 & \textbf{64.8}         & \textbf{60.7}         & \textbf{44.1} \\
    \bottomrule
    \end{tabular}%
    \vspace{-1mm}
    \caption{Compare with SOTA models trained with either the entire or a part ($x\%$) of training examples. Full means training with annotated videos.}
    \label{tab:st_img}
    \vspace{-3mm}
\end{table}

\subsection{Qualitative Results}
We visualize some VIS results obtained by SOTA offline\cite{vita} and online\cite{idol} approaches in Figure~\ref{fig:video}. The left example includes heavy occulusion caused by moving pedestrian, the swap of instance positions, and target-disappearing-reappearing. Under such case, VITA \cite{vita} fails to segment and track the pedestrian. IDOL\cite{idol} mistakenly assigns the ID of the dog in the two rightmost images, and the squatting person is recognized as a dog. In comparison, our proposed CTVIS is able to segment, classify and track all instances successfully.
For the right example, both VITA and IDOL fail to track the fish, and their ID switched after the video suddenly darkened. CTVIS also undergoes and ID switch (the middle image). Thanks to the noise introduced during training, CTVIS is more robust to tackle such occasional failure, and it reidentifies the fish later (the rightmost image).

%% file: sections/05_conclusion.tex
\section{Conclusion}
\label{sec:conclusion}
We have presented CTVIS, a simple yet effective training strategy for VIS. CTVIS aligns the training and inference pipelines in terms of constructing contrastive items. Its ingredients include long-video training, memory bank, MA embedding and noise to facilitate the learning of better instance representations, which in turn offers more stable tracking of instances. Thanks to this design, CTVIS has demonstrated superior performance on multiple benchmarks. Additionally, to relieve the cost of the video-level annotation of masks, we propose to create pseudo videos for VIS training based on goal-oriented data augmentation. CTVIS models trained with pseudo videos, which are produced using only 10\% frames extracted from the genuine training videos, achieve comparable performance, compared with SOTA models trained with full supervision. 

\textbf{Acknowledgement:}
This work was supported by National Key R\&D Program of China (No.\ 2022ZD0118700), National Natural Science Foundation of China (No.\ 62272395), Zhejiang Provincial Natural Science Foundation of China (No.\ LY21F020024), and Qin Chuangyuan Innovation and Entrepreneurship Talent Project (No.\ QCYRCXM-2022-359).